# aUToPath: Unified Planning and Control for Autonomous Vehicles in Urban Environments Using Hybrid Lattice and Free-Space Search

Tanmay P. Patel*, Connor Wilson*, Ellina R. Zhang*, Morgan Tran, Chang Keun Paik,
Steven L. Waslander, Timothy D. Barfoot
*Institute for Aerospace Studies*
*University of Toronto*
Toronto, Canada
{tanmay.patel, connor.wilson, ellina.zhang}@mail.utoronto.ca

*Abstract*—This paper presents *aUToPath*, a unified online framework for global path-planning and control to address the challenge of autonomous navigation in cluttered urban environments. A key component of our framework is a novel hybrid planner that combines pre-computed lattice maps with dynamic free-space sampling to efficiently generate optimal driveable corridors in cluttered scenarios. Our system also features sequential convex programming (SCP)-based model predictive control (MPC) to refine the corridors into smooth, dynamically consistent trajectories. A single optimization problem is used to both generate a trajectory and its corresponding control commands; this addresses limitations of decoupled approaches by guaranteeing a safe and feasible path. Simulation results of the novel planner on randomly generated obstacle-rich scenarios demonstrate the success rate of a free-space Adaptively Informed Trees* (AIT*)-based planner, and runtimes comparable to a lattice-based planner. Real-world experiments of the full system on a Chevrolet Bolt EUV further validate performance in dense obstacle fields, demonstrating no violations of traffic, kinematic, or vehicle constraints, and a 100% success rate across eight trials.

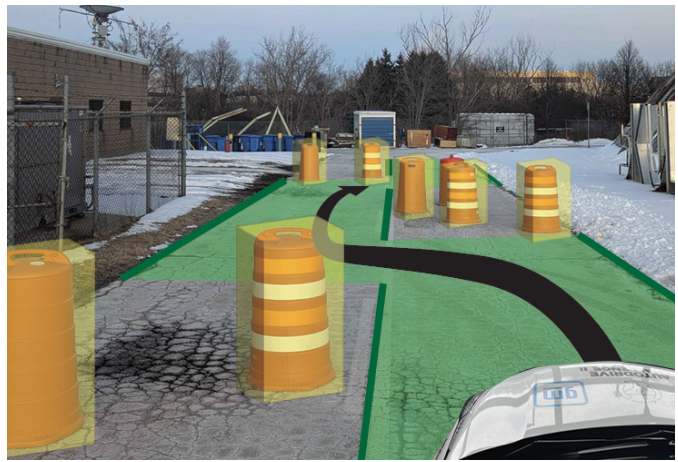

Fig. 1: The Chevrolet Bolt EUV 2022 about to drive autonomously through a dense arrangement of barrels. The vehicle will use the architecture proposed in this paper to navigate the obstructions and travel to the destination. The bounded green region denotes the 'safe' corridor available for the vehicle. The black line represents the final optimized trajectory that the vehicle will track.

## I. INTRODUCTION

Autonomous vehicles have the capacity to significantly improve the safety and reliability of our transportation systems. To achieve this, efficient motion planning and control are critical, but this can be challenging in dense urban environments with varying types, densities, and arrangements of obstructions that can change over time [1]. In particular, obtaining a feasible, high-fidelity global path for an autonomous vehicle using both a complete and efficient algorithm remains a significant challenge [2].

This paper proposes aUToPath, an architecture that leverages known maps to improve the tractability of an efficient and complete global search. aUToPath also features a joint optimization phase to transform the search result into a feasible trajectory and a set of control inputs.

This work is validated in the context of the SAE AutoDrive Challenge II, a collegiate-level event hosted by SAE International and General Motors [3]. The competition invites design teams from North American universities to participate annually in a week-long competition where they are assessed on their ability to complete self-driving scenarios. The University of Toronto has competed successfully since 2018 [4] and placed first in all events at the 2024 competition; aUToPath is a novel improvement to a major component of our competition architecture.

Owing to computational limitations, we focus on lightweight, resource-efficient path-planning methods. Unlike some traditional approaches [5], our planner leverages high-fidelity, pre-built maps covering the entire driveable area. These maps are condensed and geographically bounded, enabling online search through them [3]. The planner can operate within this pre-defined environment and only respond dynamically to obstacles while adhering to traffic laws, such as signals and signs. This insight significantly alleviates

---

*These authors contributed equally to this work.

computational load.

Our key contribution lies in combining a lattice planner [6]—derived from the known map—with selective and localized free-space sampling. This maintains the computational efficiency of a lattice search while introducing the asymptotic optimality of sampling-based methods [7, 8], enabling fast and robust global planning.

We guarantee feasibility by constructing safe corridors around the search result—bounded regions within which the vehicle can safely operate. We then use model predictive control (MPC) based on sequential convex programming (SCP) [9] to both generate and execute a kinematically feasible, collison-free trajectory. In-silico and in-car experiments validate the practical utility of our hybrid approach on a wide range of dense urban driving scenarios.

The remainder of this paper is organized as follows: Section II reviews related work in motion planning and control for autonomous vehicles. Section III details the proposed methodology, including the hybrid planner, the corridor generator, and the SCP-based MPC. Section IV presents results from closed-loop simulations, while Section V discusses in-car behaviour. Finally, Section VI concludes with a summary and directions for future work.

## II. Related Work

### A. Motion Planning in Self-Driving Vehicles

A common approach to motion planning in urban environments is the use of graph-search methods [5], such as a lattice-based planner [6]. This method pre-computes a graph of feasible splines, including all driveable lanes and lane changes (see Fig. 2 for an example). During the path-planning stage, a graph-search algorithm is used to generate a continuous reference path. While this online computation can be performed rapidly, lattice-based methods can suffer from the following shortcomings:

1) the lattice planner is not resolution complete, meaning it may fail to find a feasible path through a cluttered environment even if one exists;
2) the lattice discretization is performed without any knowledge of obstacle locations, which can lead to a highly sub-optimal solution even when one is found, particularly in narrow or dense passages that the planner did not originally recognize as such; and
3) a path can only be generated from the pre-computed lattice edges, which can make it difficult or impossible to recover if the vehicle significantly deviates from the lattice due to disturbances.

The fundamental issue with lattice planners is their *a prori* discretization of the space [11], which limits the number of cases that can be handled, as observed in Section V.

Another common method for motion planning in autonomous vehicles is using sampling-based approaches, such as Rapidly-Exploring Random Trees* (RRT*) [12, 13] or its variants, notably Batch-Informed Trees* (BIT*) [8], which is almost-surely asymptotically optimal [7]. These methods

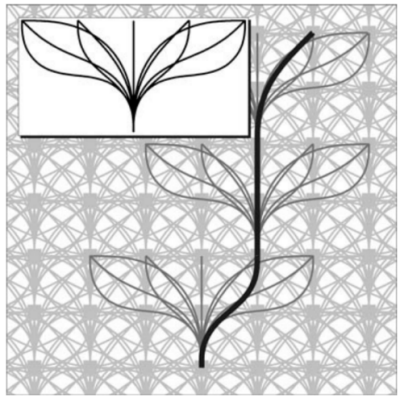

Fig. 2: An example of a lattice search space. This plot, originally from [10], depicts a series of parallel lanes, as well as all possible lane changes, in a pre-computed graph (the light grey lines). Each edge can be given a cost and a search algorithm can be used to generate a reference path, such as the black line. The dark grey segments denote edges that were expanded during the search but were not part of the solution.

randomly sample states and connect them when appropriate, creating a graph in the state space.

An example of using sampling-based approaches for practical settings is provided by [14], which introduces the idea of using a lower-fidelity BIT*-based planner to identify the homotopy class of the solution. This homotopy class, combined with a graph of kinematically feasible, pre-defined references, enables the creation of a high-fidelity global reference path. While we do not have access to pre-defined feasible paths, we can take advantage of lane centrelines when available as pseudo-references after the appropriate homotopy class has been identified. However, when performing lane changes or navigating dense obstacle fields, we lose the notion of a pre-defined reference.

Finally, there is increasing interest in neural planning and end-to-end machine learning for self-driving [15]. However, these approaches are not suitable owing to computational constraints imposed by our platform (see Section V).

### B. Adaptively Informed Trees* (AIT*)

Adaptively Informed Trees* (AIT*) [7] is a sampling-based motion planning algorithm that combines the advantages of heuristic-informed search and lazy edge evaluation. It extends BIT* by introducing an *adaptive, asymmetric bidirectional search* for fast and asymptotically optimal path planning, particularly in domains with expensive edge evaluations and large search spaces. This makes it suitable for our application.

*a) Informed Sampling and Edge-Implicit Graphs:* Like BIT*, AIT* incrementally builds a random geometric graph (RGG) over batches of samples using informed sampling. This focuses sampling within an *informed set*, i.e., the subset of the state space that could potentially yield lower-cost paths than the current best. The RGG edges are processed in order of increasing estimated solution cost, based on a composite heuristic $f(x) = g(x) + h(x)$, where $g(x)$ is the cost-to-come and $h(x)$ is the heuristic cost-to-go.

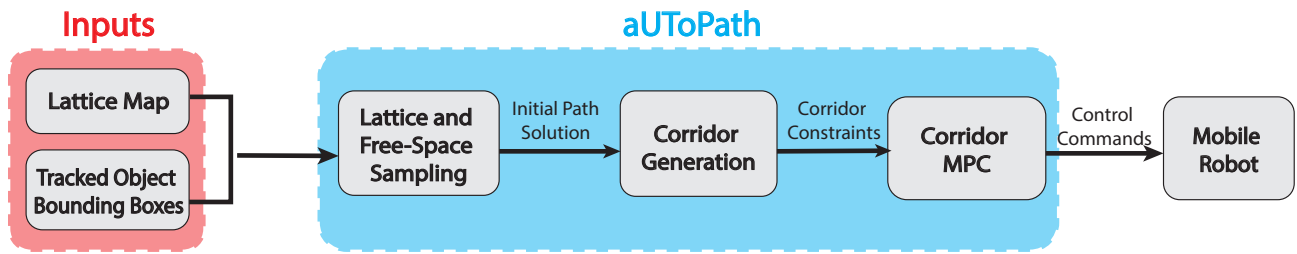

Fig. 3: A high-level overview of the proposed system architecture.

*b) Asymmetric Bidirectional Search:* AIT* augments BIT* with an adaptive reverse search. The forward search follows BIT*, but its heuristic $h(x)$ is not static——it's computed on-the-fly using a *lazy reverse search* from the goal. This search is implemented with Lifelong Planning A* (LPA*), which incrementally refines the heuristic values based on edge evaluation outcomes and the structure of the sampled graph.

*c) Adaptive Heuristic Estimation:* Instead of relying on a fixed heuristic, AIT* continuously improves its heuristic estimate through experience. When the forward search detects an invalid edge, it updates the reverse tree, which repairs the heuristic using LPA*. This creates a feedback loop: the forward search evaluates real edges and informs the reverse search of obstacles, while the reverse search provides increasingly accurate heuristics to the forward search.

Overall, AIT* preserves the asymptotic optimality of BIT* while improving performance in practice for domains with expensive edge costs, making it a promising avenue.

### C. Control Systems in Self-Driving Vehicles

Self-driving vehicles increasingly use MPC to plan optimal control commands for fixed and/or receding horizons. For instance, MPC can be used to directly track the result of a path generated by a sampling-based approach [16]. This, however, assumes that the path is kinematically feasible and offers no safety guarantees otherwise. Our work expands the responsibility of the MPC such that it can behave simultaneously as a local path-planner and a controller, as proposed by Liu et al. [17]. This addresses the issue of feasibility since the MPC can generate a trackable path directly when a reference is not available.

To improve runtime when solving MPC optimizations that would otherwise be non-convex, SCP can be used to iteratively approximate the problem. It generates solutions comparable to a standard MPC, while reducing the computational burden [9].

### D. Combined Motion Planning and Control Systems

Previous works have proposed unified motion planning and control, such as in autonomous racing [18]. Their work uses model predictive contouring control to jointly optimize for a local path and control commands along a pre-defined racetrack under limited time constraints. However, the work relies on a continuous *de facto* reference path (the centreline) about which to formulate the optimization problem, similar to many others [14, 18, 19, 20]. We do not always have access to such a path, as outlined in Section II-A.

However, we are not the first to propose joint planning and control in the absence of a pre-defined reference path. Liu et al. [17] integrate path-planning and tracking into a unified framework. Their approach leverages *a priori* road information, such as available lanes and road boundaries, to compute driveable lanes and control commands in a single optimization problem. However, they use a highly simplified unicycle model to represent their simulated vehicle and do not assess their system on a physical platform, since their primary objective is to demonstrate the usage of MPC for simulated behaviour planning (i.e., deciding mode switches like lane changes or highway ramp merging). In contrast, our architecture prioritizes practical deployment and control on a physical vehicle, where runtime and reliability are critical.

## III. METHOD

Our system is composed of three modules—path-planning, corridor generation, and corridor MPC—as outlined in Fig. 3.

### A. Path-Planning: A Hybrid Lattice and Free-Space Strategy

The path-planning module uses an extended version of AIT* [7] that integrates the known lattice map with free-space samples. Our usage extends out-of-the-box AIT* in the following ways:

1) the search is performed only on the existing road map during the first iteration of the algorithm;
2) during subsequent iterations, we begin sampling the free-space in regions informed by previous searches; and
3) a reverse tree is built on initialization and maintained throughout the lifetime of the planner.

During the first iteration of AIT*, we only search precomputed map points, which lie along lane centrelines. Note that we can pre-compute the AIT* heuristic offline across the full map to ease online computation, taking advantage of known relationships and distances between map nodes. This initial search either results in a rudimentary lattice solution if one exists or fails otherwise. In either case, we keep track of all obstacles observed during this iteration.

During subsequent iterations, we introduce random free-space samples near obstacles that were observed during previous iterations (see Fig. 4). This is achieved by sampling poses from multivariate Gaussian distributions centered around the map points closest to where obstacles were encountered. In addition to these new samples, we continue to leverage map relations during these iterations.

The first iteration of our algorithm exclusively searches the lattice, exploring all possible lattice routes in order of optimality. This ensures that if no solution is found initially, the lattice has been thoroughly searched for feasible paths. In subsequent iterations, the algorithm incorporates local free-space samples guided by our selective sampling strategy. This biases sampling toward regions where an improvement over the original lattice solution is most likely to be found.

Although biased towards promising regions identified during previous searches, our adaptation retains an underlying uniform distribution of samples. This means it is almost-surely asymptotically optimal [8], unlike a lattice-only search.

*Algorithm Summary:*
1) Build a reverse tree rooted at the destination by computing optimal costs-to-go without considering collisions.
2) Perform a forward search from the vehicle pose using the cost-to-go values from the reverse tree as the heuristic. When an edge is in collision with an obstacle, remove it from the reverse tree and update cost-to-go values.
3) Using the previous search, identify nodes in collision with obstacles and sample a Gaussian distribution centered at those nodes. Connect these samples to the existing graph using a nearest-neighbour search and update the reverse tree accordingly.
4) Repeat the procedure from the forward search.

A general pseudocode for our method, including our extensions to standard AIT*, is provided in Algorithm 1. A more specific overview can be seen in Algorithm 2, with supporting procedures outlined in Algorithm 3.

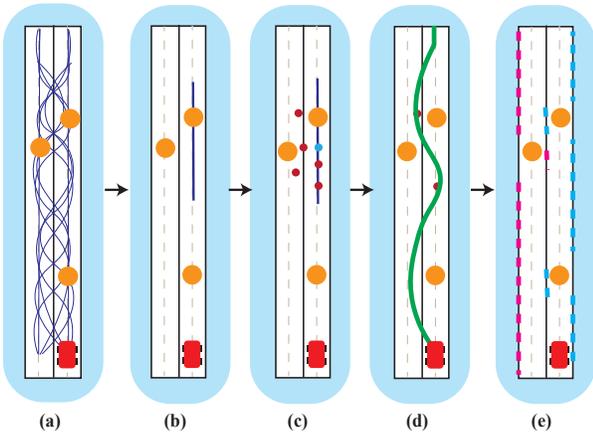

Fig. 4: Our method for adding free-space samples during path planning. The obstacles are in orange and the vehicle is in red. (a) The lattice planner is unable to find a path. (b) The planner identifies regions for sampling, guided by the closest attempt of the lattice planner to find a path. (c) Free-space samples are generated around these map points. (d) Samples are added during the graph search, and a feasible path is found. (e) The corridor is constructed as per Section III-B. For free-space points that were introduced into the path, the associated road/lane boundaries are determined from the map point around which the free-space point was sampled.

**Algorithm 1** Outline of the planning algorithm with changes to AIT* identified in blue

Create a graph using the existing road map.
Build a reverse tree by computing cost-to-go values of nodes in the graph.
Initialize a queue with edges which connect the vehicle to nodes in the graph.
**while** *the timeout has not been reached* **do**
    Select the edge from the queue with the minimum cost using the cost-to-go heuristic from the reverse tree.
    **if** *the edge may improve the solution* **then**
        **if** *the edge is collision-free* **then**
            Add the edge to the forward tree.
        **else**
            Save the nodes in the edge for sampling after the search iteration completes.
            Remove the edge from the reverse tree and compute updated cost-to-go values.
        **end**
    **else**
        Sample Gaussian distributions centered at nodes identified in the previous search.
        Sample a uniform distribution over the free space.
        Connect samples to the existing graph.
        Update the reverse tree.
        Re-initialize the queue.
    **end**
**end**

**Algorithm 2** Overview of core planning procedures

**Input:** Map data, current pose, destination, config
**Output:** Path from current pose to destination
**Procedure** *InitializeMap()*
    Load map nodes, edges; build R-tree index; connect nodes
**Procedure** *InitializeDestination(dest)*
    Reset all $node.dist\_to\_goal \leftarrow \infty$; mark dest. nodes with $dist\_to\_goal = 0$; propagate goal distances
**Procedure** *FindPath()*
    Clear search data; connect vehicle pose to roadmap; initialize queue
    **while** $time\_elapsed < max\_planning\_time$ **do**
        $obstacle\_encountered \leftarrow False$
        **while** *queue not empty* **do**
            Check next best edge with A* heuristic to get $path$
            **if** *path cut-off by obstacle* **then**
                $obstacle\_encountered \leftarrow True$
            **end**
            **else if** *found a solution* **then**
                $path \leftarrow$ most recent solution
            **end**
        **end**
        **if** $obstacle\_encountered$ **then**
            SAMPLE()    // **Focused Sampling**
        **end**
    **end**
    **return** $path$

## Algorithm 3 Supporting procedures

**Procedure** *Sample(num_samples)*
  **for** $i = 1$ **to** *num_samples* **do**
    Select random *node* from *invalid_nodes*
    *new_pose* ← SAMPLEGAUSSIAN(*node.pose*, $\sigma_{long}$, $\sigma_{lat}$, $\sigma_{yaw}$)
    **if** *collision-free(new_pose)* **then**
      Create *new_node*
      CONNECTSAMPLE(*new_node*) // **Map-Aware Node Connections**
      Add new node to queue
    **end**
  **end**

**Procedure** *ConnectSample(node)*
  **for** *each nearby node in current graph* **do**
    **if** *distance, heading changes within thresholds* **then**
      **if** *maneuver is legal according to map* **then**
        Create spline connection
      **end**
    **end**
  **end**

### B. Corridor Generation

The objective of this module is to convert the path computed by the previous module into a set of reasonable boundaries that the vehicle can operate within, known as the *corridor*. The most intuitive way to achieve this is to inflate the path laterally in both directions to the minimum of the following values:

1) the lateral distance to the nearest obstacle; and
2) the lateral distance to the nearest lane boundary if we are following the centreline, or to the boundaries of adjacent lanes if we are changing lanes/avoiding obstacles (this generalizes to an arbitrary number of lanes).

A key insight is that in practice, explicitly computing the algebraic representation of the corridor boundaries is unnecessary and introduces additional complexity. Instead, the corridor serves as a framework for defining the homotopy class within which the path optimization occurs [14].

The lane and road boundaries are naturally represented as a sequence of points, while obstacles can be expressed as the corner points of their polygons. Rather than joining these points to generate piecewise continuous functions for the boundaries, we interact directly with the points themselves. At each optimization step, we determine whether each boundary or obstacle point lies to the left or right of the vehicle as it traverses the path. This is achieved by evaluating the position of each point relative to the hybrid planner's current solution.

Using this positional information, we apply linear constraints to the vehicle's trajectory within the MPC horizon, computed from the boundary points on either side. These constraints, described in (6) and (7), ensure that the vehicle remains within the defined corridor. Fig. 5 illustrates the process of corridor generation.

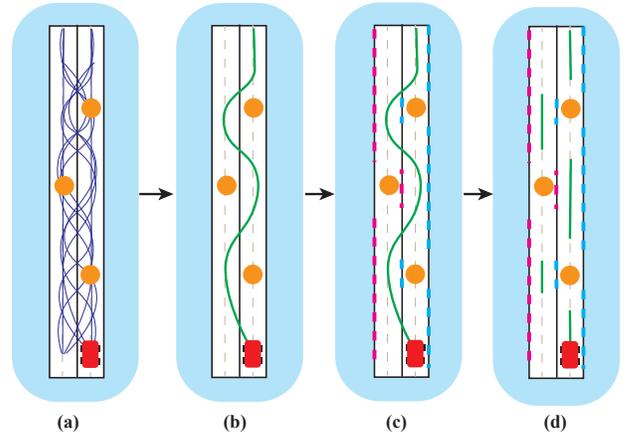

Fig. 5: Our method for constructing the corridor of driveable space, which is ultimately used as a positional constraint for the MPC. The obstacles are in orange and the vehicle is in red. (a) We search the lattice to find a feasible path with no collisions. (b) A feasible path is found. Free-space samples are added (see Fig. 4) and are used to find a lower-cost path. (c) For each point in the feasible path, the closest boundary points are identified. These boundary points are either road boundaries or the edge of an obstacle, depending on which point is closer to the path. (d) The boundary points serve as positional constraints for the MPC detailed in Section III-C, and lane centrelines—if and where available—are used as the reference path. Note that the Figure uses magenta versus blue boundary points to distinguish whether they should fall on the left or the right, respectively, of the vehicle at the time of passing.

### C. Corridor Model Predictive Control

The corridor MPC—which draws from [9] and is formulated using SCP—is the final module in our system and is intended to optimize for both a local trajectory and its associated control commands such that the vehicle remains within the corridor computed by the previous module. The state vector is

$$\mathbf{x} = \begin{bmatrix} x & y & \theta & v & \psi \end{bmatrix}^\mathsf{T}, \tag{1}$$

where $x$, $y$, $\theta$ denote the vehicle's pose, $v$ denotes longitudinal speed and $\psi$ denotes the steering angle—the angle between the vehicle's front wheels and its longitudinal axis. The control input is

$$\mathbf{u} = \begin{bmatrix} a & \Delta\psi \end{bmatrix}^\mathsf{T}, \tag{2}$$

where $a$ denotes longitudinal acceleration and $\Delta\psi$ denotes the steering command, i.e., change in steering angle. As per typical model predictive control, a full sequence of optimal control commands is computed at each timestep but only the first is applied.

The discretized state model, imposed as a constraint, is

$$\mathbf{x}_{k+1} = \begin{bmatrix} x_k + d\cos(\psi_k + \Delta\psi_k) \\ y_k + d\sin(\psi_k + \Delta\psi_k) \\ \theta_k + d\tan(\psi_k + \Delta\psi_k)/l \\ \sqrt{v_k^2 + 2a_k d} \\ \psi_k + \Delta\psi_k \end{bmatrix}, \tag{3}$$

where $d$ is a fixed distance between consecutive points in the solution path, and $l$ is the distance from the rear axle to the vehicle's centre of mass. The optimization problem, expressed in the map frame, is

$$\min_{\mathbf{u}} \sum_{k=1}^{N} w_1(v_k - v_{k,\text{ref}})^2 + w_2\psi_k^2 + w_3 a^2 + w_4(\Delta\psi)^2$$
$$+ w_{k,5}((x_k - x_{k,\text{ref}})^2 + (y_k - y_{k,\text{ref}})^2)$$
$$+ w_6 \max(\sigma_{k,\text{l}}, 0) + w_7 \max(\sigma_{k,\text{r}}, 0) \quad (4)$$

where

$$w_{k,5} = \begin{cases} \tilde{w}_5, & \text{reference path exists at step } k \\ 0, & \text{no reference path exists at step } k \end{cases} \quad (5)$$

subject to

$$\alpha_{k,\text{l}} x_k + \beta_{k,\text{l}} y_k \leq \gamma_{k,\text{l}} + \sigma_{k,\text{l}}, \quad \sigma_{k,\text{l}} \leq \sigma_{\text{buffer}} \quad (6)$$
$$\alpha_{k,\text{r}} x_k + \beta_{k,\text{r}} y_k \leq \gamma_{k,\text{r}} + \sigma_{k,\text{r}}, \quad \sigma_{k,\text{r}} \leq \sigma_{\text{buffer}} \quad (7)$$
$$0 \leq v \leq v_{\max} \quad (8)$$
$$-\psi_{\max} \leq \psi \leq \psi_{\max} \quad (9)$$
$$-a_{\max} \leq a \leq a_{\max} \quad (10)$$
$$-\Delta\psi_{\max} \leq \Delta\psi \leq \Delta\psi_{\max}, \quad (11)$$

where $N$ is the prediction horizon, the subscript 'ref' denotes reference values, $\sigma_{k,\text{l}}$ and $\sigma_{k,\text{r}}$ are slack variables, and $w_1$ through $w_7$ are costs. Note that $w_{k,5}$ is the tracking cost and is designed to allow us to benefit from a convenient reference path when it exists (e.g., when driving in a single lane), and plan from scratch otherwise (e.g., when navigating around a roadway obstruction, or performing a lane change).

The two linearized corridor constraints, informed by the corridor generation module, are denoted by (6) and (7), where $\alpha$, $\beta$, and $\gamma$ are standard form coefficients of a line. The subscript 'l' indicates the left constraint and the subscript 'r' indicates the right constraint. Each one consists of a line inflated outward from the vehicle's pose to the nearest lane/road boundary or obstruction on that side (from the previous module), with a safety buffer of $\sigma_{\text{buffer}}$ subtracted.

The slack variables, $\sigma_{k,\text{l}}$ and $\sigma_{k,\text{r}}$, are intended to penalize the vehicle for entering the safety buffer located along the outer edges of the corridor, i.e., they do not contribute to the cost unless the vehicle is within $\sigma_{\text{buffer}}$ of the corridor edge.

The use of SCP for this problem arises from the need to iteratively re-compute a linearized corridor constraint based on the most recent solution path. A single optimization pass would linearize the corridor constraint only once, yielding a worse approximation of the true feasible region (this is empirically validated in Section IV-A).

Initially, the corridor is linearized with respect to the raw planner solution—specifically, the segment closest to the starting point of the proposed path. However, as the optimization progresses, the updated solution path may align more closely with a different segment of the corridor. In such cases, the corridor must be re-linearized relative to this new segment to better reflect the environment's actual physical constraints.

By iteratively re-linearizing and solving the optimization problem, SCP ensures that the solution gradually converges towards a more accurate and feasible path. Iterative refinement addresses the challenges posed by poor initialization and dynamically adjusts the corridor to account for changes in the solution geometry.

Another benefit of SCP is that it explicitly offers a tradeoff between solution quality and computational effort. The number of iterations, $k$, can be chosen empirically to balance these two attributes as desired. This is carried out for our system in Section IV-A.

## IV. Simulation Experiments

An internally developed closed-loop simulator is used to study the performance of our system. The simulator employs the dynamic bicycle model and uses the same weight, axle-to-centre-of-mass distance, and steering ratio as the Chevrolet Bolt EUV 2022, our physical platform. The simulator allows the user to create scenarios at real-world locations with common roadway items and obstructions, which include barrels, barricades, signs, traffic lights, pedestrians, etc.

### A. Study of SCP Iterations and Solution Quality

The focus of this section is to assess our formulation of the corridor MPC as an SCP problem. The purpose of SCP is to enable higher-fidelity approximations of the physical corridor (see Section III-C), which should improve safety. We define safety in terms of the minimum distance to an obstacle, with larger distances being considered safer. This is tested over twenty-five simulated scenarios, each repeated five times for different SCP iteration counts. A baseline of one SCP iteration is included, in which the MPC uses the planned path directly without further iteration. To create test cases, we define a *root* scenario with four barrels. For each generated case, we apply a random perturbation ($\sim \mathcal{N}(0,1)$ metres) to each barrel's location while ensuring the barrels remain on the road.

We compare the safety of different vehicle trajectories by first measuring the minimum distance between the vehicle position and any surrounding obstacle across the full trajectory, then ranking the five different SCP configurations in descending order by this metric. We expect safer solutions with more iterations due to a stronger approximation of the corridor and therefore, greater utility of the slack costs introduced in (4).

Fig. 6 illustrates that as the number of iterations increases, the prevalence of safer solutions (depicted in green) rises. Additional SCP iterations increase the minimum distance between the solution path and surrounding obstacles, demonstrating the effectiveness of SCP in improving safety margins. There is one notable exception where the largest safety margin occurs with the fewest number of iterations. In this randomly generated scenario, the barrels were widely spaced, leading to very similar solutions across SCP configurations. The difference in the minimum distance to a barrel between the one-iteration and five-iteration cases was just 6 cm.

We also study the effect of varying the number of SCP iterations on the optimization time. As expected, Fig. 7 shows an increase in optimization time as the number of SCP iterations is increased.

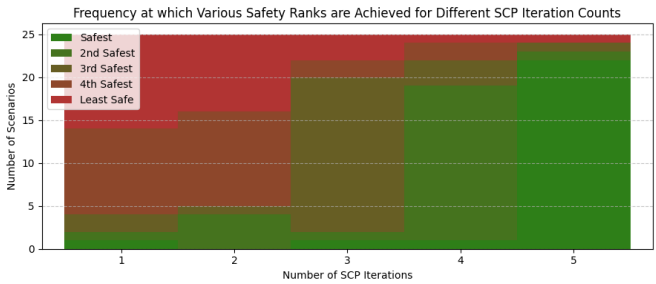

Fig. 6: An illustration of the safety rank achieved by different SCP iteration counts over twenty-five simulated scenarios. Green indicates that a particular scenario was completed most safely (largest distance to obstacles when passing) while red indicates it was completed least safely.

Guided by these results, we elect to run aUToPath with four iterations of SCP. While Fig. 6 highlights an improvement in safety with additional iterations, the advantage is marginal between the fourth and fifth iterations. This is compounded by a significant increase in computation cost, as per Fig. 7. Therefore, four SCP iterations is observed to empirically strike a balance between performance and practicality, and is used for all further experiments.

### B. Study of the aUToPath Hybrid Planner

This section assesses the performance of the novel hybrid planner from Section III-A. Our proposed method aims to offer a compromise between success rate and search time during a large global search. Since our method is a hybrid of free-space and lattice-based search, each of the respective modes naturally serve as baselines. The free-space baseline uses out-of-the-box AIT* to enable a direct comparison to our method.

For this experiment, we generate two scenarios. In scenario A, four roadway items are placed 7 metres apart. In scenario B, two obstructions are placed just 2 metres apart, creating a narrow passageway only slightly wider than the vehicle. Both scenarios are depicted in Fig. 8. In both cases, the system is required to find a solution to a destination point approximately 180 metres from the initial location. Each scenario is repeated twenty times when assessing the two sampling-based approaches, but only once when assessing the lattice baseline since it behaves deterministically.

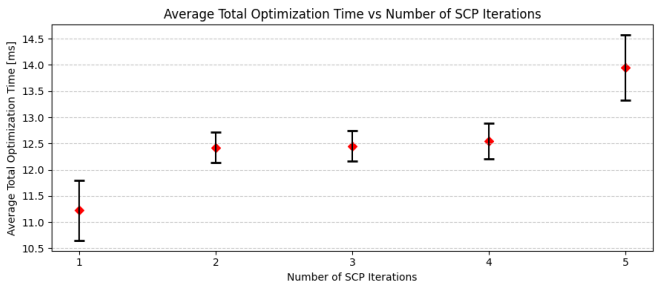

Fig. 7: Average time across trials to perform all SCP iterations for various iteration counts. The error bars depict the standard deviation across trials.

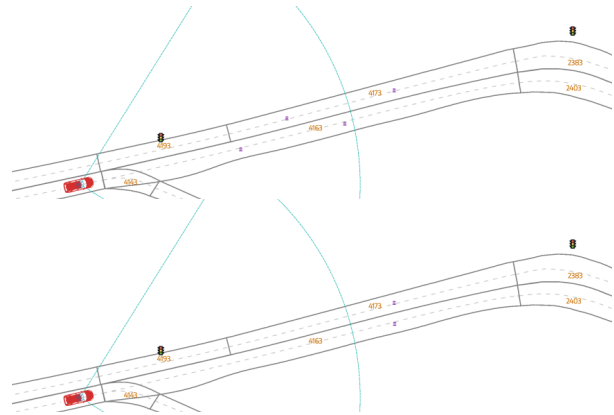

Fig. 8: A depiction of the two scenarios in our simulator. The red car represents the ego-vehicle, and the cyan cone indicates its field of view as obtained from the physical platform.

Table I compares the computation time required to find a solution for the two planning scenarios. In scenario A, a lattice solution exists and is quickly identified by both the lattice planner and aUToPath. AIT* is far slower since it does not employ prior map information. In scenario B, there is no lattice solution, and while AIT* finds a solution, aUToPath is significantly faster since it identifies and biases sampling towards promising regions of the state space by leveraging its initial lattice search. Together, these results demonstrate that our hybrid method achieves the robustness of a full free-space search without incurring the same computational penalty.

## V. EXPERIMENTS ON THE CHEVROLET BOLT EUV

The performance of our system is verified through real-world experiments on the Chevrolet Bolt EUV. Our compute platform is the M50FCP Intel server. This platform features 64 GB RAM and two Intel Xeon Gold 5420+ Processors with 56 cores, a 2.0 GHz base clock speed and a 2.7 GHz turbo boost clock speed. It also houses an Intel Data Center Flex 170 GPU, which is utilized for running the deep learning models for the perception stack. ROS2 Humble [21] was used to implement and run our system on the vehicle.

We present results on a scenario in which the vehicle is required to navigate a set of four barrels placed 6 metres apart and in alternating lanes, then proceed along a gently curved road and park in an available slot. Fig. 9 provides a birds-eye view of the course. The objective is to assess our system's ability to repeatedly complete a challenging scenario and to measure the quality of the control commands generated. The success rate (as defined in Section IV-B) is computed over eight trials and compared to a lattice baseline.

TABLE I: Comparison of the aUToPath Selective Sampling Technique to Baselines Based on Success Rate and Solution Length

| Method | Scenario A | | Scenario B | |
|---|---|---|---|---|
| | Solution Time [ms] | Length [m] | Solution Time [ms] | Length [m] |
| aUToPath | 23 | 188.921 | 2683 | 188.886 |
| Lattice | 21 | 188.921 | N/A | N/A |
| AIT* | >10000 | 190.233 | >10000 | 183.943 |

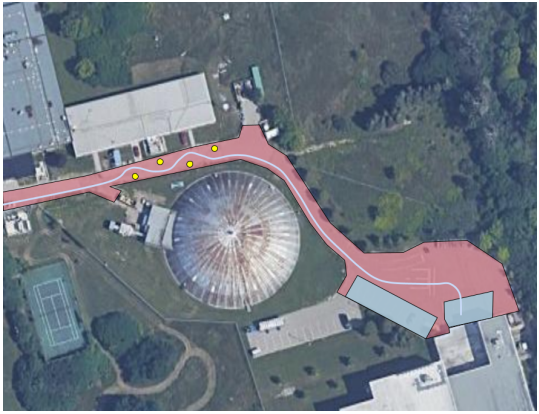

Fig. 9: A bird's eye view of the experiment setup. The red region indicates driveable space, while the blue region indicates the parking area. The yellow circles denote barrels while the light grey line indicates a potential trajectory that aUToPath might track.

Additionally, we evaluate a series of vehicle dynamics metrics, which are closely related to the vehicle states and control inputs introduced in Section III-C. Constraints to these values, imposed directly by the vehicle or indirectly for passenger comfort, are compiled in Table II, where each constraint applies to both positive and negative values. The system should ensure that these constraints are never violated.

Our system responds to all four obstacles in the scenario, constructs a corridor between them, and optimizes for a trajectory and a set of control commands that results in successful completion for all eight trials. An attempt to use the baseline lattice architecture on this scenario fails between the first and second barrel, despite successful detection of all obstacles. The repeated failure can be attributed to the dense obstacle configuration, which is more restrictive than typical scenarios.

The full trajectory of vehicle commands over a sample trial are presented in Fig. 10. Table III then presents the maximum absolute vehicle dynamics metrics observed across all eight trials, compared against the constraints from Table II. No constraint was violated at any point, demonstrating our system's consistent and safe operation.

TABLE II: The Set of Constraints Imposed on Our Vehicle

| Metric | Passenger Comfort or Vehicle Constraint |
|---|---|
| Long. Accel. [m/s$^2$] | 3.00 |
| Long. Jerk [m/s$^3$] | 0.90 |
| Lat. Accel. [m/s$^2$] | 3.00 |
| Steering Angle [rad] | 0.52 |
| Axle Torque [kN·m] | 2.00 |

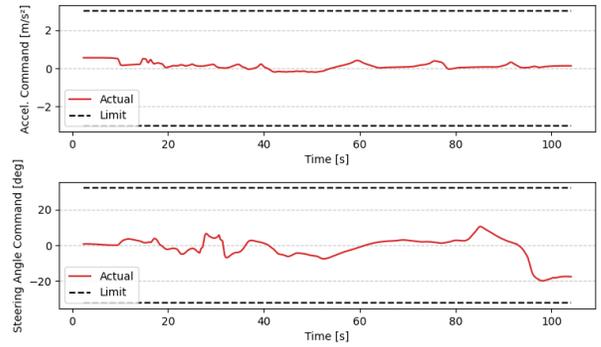

Fig. 10: Vehicle commands sent during an in-car trial, alongside vehicle constraints imposed by the platform.

TABLE III: Controller Metrics Across Trials

| Metric | Maximum Across Trials | Passenger Comfort or Vehicle Constraint |
|---|---|---|
| Long. Accel. [m/s$^2$] | 1.94 | 3.00 |
| Long. Jerk [m/s$^3$] | 0.047 | 0.90 |
| Lat. Accel. [m/s$^2$] | 2.80 | 3.00 |
| Steering Angle [rad] | 0.50 | 0.52 |
| Steering Rate [rad/s] | 0.021 | N/A |

Live footage of our system navigating this scenario can be found here.

## VI. CONCLUSION

We present aUToPath, an architecture for joint global motion planning and control in urban self-driving. Our hybrid planner combines lattice-based planning with corridor-based optimization, operating without a continuous reference path. It balances optimality and speed for real-time use, outperforming a lattice baseline in challenging scenarios while maintaining global planning under diverse obstacle configurations. Future work will focus on improving robustness to upstream errors, such as false detections and localization drift.

## ACKNOWLEDGMENT

The authors thank General Motors and SAE International for sponsoring the SAE AutoDrive Challenge II, and the University of Toronto Faculty of Applied Science and Engineering Dean's Strategic Fund for financial support. They also thank Jayce Wang for contributions to early development, and Professor Gabriele M.T. d'Eleuterio for valuable feedback.